\documentstyle[12pt,epsf]{article}

\newcommand{\bm}[1]{\mbox{\boldmath $#1$}}
\newcommand{\be}{\begin{equation}}
\newcommand{\ee}{\end{equation}}
\newcommand{\bea}{\begin{eqnarray}}
\newcommand{\eea}{\end{eqnarray}}
\newcommand{\nn}{\nonumber}
\newcommand{\ov}{\overline}

\makeatletter
\def\section{\@startsection{section}{1}{\z@}{-3.5ex plus -1ex minus -.2ex}{2.3ex plus .2ex}{\normalsize \bf}}
\def\subsection{\@startsection{subsection}{2}{\z@}{-3.25ex plus -1ex minus -.2ex}{1.5ex plus .2ex}{\normalsize \sl}}
\def\subsubsection{\@startsection{subsubsection}{3}{\z@}{-3.25ex plus -1ex minus -.2ex}{1.5ex plus .2ex}{\normalsize}}
\makeatother

\makeatletter
\def\@evenhead{
	\vbox{\hbox to\hsize{\bf \thepage \hfill \sl \evenmark}
	}
}
\def\@oddhead{
	\vbox{
		\hbox to\hsize{\oddmarkA \oddmarkB \hfill \oddmarkC}
		\hbox to\hsize{\hfill \oddmarkD}
		\vspace{.03in}
		\hbox to\hsize{\hfill \oddmarkE}
		\vspace{.03in}
		\hbox to\hsize{\hfill \oddmarkF}
	}
}
\def\@evenfoot{\vbox{\hbox to\hsize{\hrulefill}
	\vspace{-40pt} \hbox to \hsize{\today \sl \footmsg \hfill}
	}
}
\def\@oddfoot{
	\vspace{-40pt} 
	\hbox to \hsize{ \footmsgA \hfill }
}
\makeatother

\def\evenmark{\bf entropy}
\def\oddmarkA{{\sl A Bayesian Reflection on Surfaces}}
\def\oddmarkB{\thepagerange}
\def\oddmarkC{}
\def\oddmarkD{}
\def\oddmarkE{}
\def\oddmarkF{}
\def\footmsgA{{\copyright 1999 by the author. Reproduction for noncommercial purposes permitted.}}

\def\thedoctitle{\bf A Bayesian Reflection on Surfaces}

\def\theauthorname{David R. Wolf}

\def\authoraddress{PO 8308, Austin, TX 78713-8308, USA \\ E-mail: drwolf@realtime.net}

\def\therevisionhistory{Revision history: 1st, January 1998, 2nd, June 1998, 3rd, October 1999}

\begin{document}

{
	\noindent {\\ \\}
}

{
	\LARGE \noindent \thedoctitle
} \\

{
	\noindent {\bf {\theauthorname}}
} \\

{
	\normalsize \noindent {\authoraddress}
} \\

{
	\noindent {\sl \therevisionhistory}
} \\

\vspace{2pt} \hbox to \hsize{\hrulefill}
\vspace{.1in}

\noindent
{\bf Abstract:} The topic of this paper is a novel Bayesian continuous-basis field representation and inference framework.  Within this paper several problems are solved:  The maximally informative inference of continuous-basis fields, that is where the basis for the field is itself a continuous object and not representable in a finite manner;  the tradeoff between accuracy of representation in terms of information learned, and memory or storage capacity in bits;  the approximation of probability distributions so that a maximal amount of information about the object being inferred is preserved;  an information theoretic justification for multigrid methodology.  The maximally informative field inference framework is described in full generality and denoted the Generalized Kalman Filter. The Generalized Kalman Filter allows the update of field knowledge from previous knowledge at any scale, and new data, to new knowledge at any other scale. An application example instance, the inference of continuous surfaces from measurements (for example, camera image data), is presented. \\


\vspace{2pt} \hbox to \hsize{\hrulefill}

\newpage


\def\oddmarkC{\thepage}
\def\oddmarkB{}
\def\oddmarkD{}
\def\oddmarkE{}
\def\oddmarkF{}
\def\footmsgA{}


\section{Overview}

The paper begins by reviewing traditional approaches to surface representation and inference.  Then the new field representation and inference paradigm is introduced within the context of maximally informative (MI) inference~\cite{WOLF99}, early ideas appearing in~\cite{WOLF96}.  The knowledge representation distribution is introduced and discussed in the context of MI inference.  Then, using the MI inference approach, the here-named Generalized Kalman Filter (GKF) equations are derived for a specific example instance of inferring a surface height field.    The GKF equations motivate a location-dependent adaptive scale or multigrid approach to the MI inference of continuous-basis fields.

\section{Introduction: Surface representation}

\subsection{Traditional methods}

Many methods for representing surfaces have been utilized previously, however these methods involve representing the surface by a {\em discrete} basis field, perhaps with a deterministic interpolation defined (bi-linear, tensor B-splines, etc.) to provide a definition for the surface at points 
intermediate to the discrete field. Probability distributions or densities of 
these discrete fields then often take the form of normalized exponentials of sums of clique energy functions, and produce a construct commonly
known as a Markov Random Field. (See Geman~\cite{GEMA84}, for an often cited example.)
There are several immediate observations on these approaches:
\begin{itemize}
\item
The surface remains unspecified at points intermediate to the
discrete field, except by the often undefined notion of interpolation.
\item
When interpolation is {\em not} defined, the discrete field probability
distribution says nothing about the probability distribution of surface
at points intermediate to the discrete field points.
\item
When interpolation is defined then, given a value of the discrete field, there is {\em no uncertainty} in the surface intermediate to the
discrete field points.  There is a deterministic mapping from any
given discrete field to the corresponding continuous surface.
In particular, when the discrete field basis covers a fixed grid on the $(x,y)$ plane with $z$ heights at each grid point, known here as a height field, all sampling of the surface intermediate to the fixed grid is determined at the scale of the fixed grid.  This is generally not physical, see next.
\item
The surface distribution is not an intrinsic property of any
physical surface, rather a post-hoc imposition of the
analyst attempting a useful regularization. For instance, 
necessary scaling properties are ignored:  Moving a camera closer to the
surface, for example, so that the density of sample points on the physical surface increases, is not properly represented in the fixed basis of the discrete field distribution;  there is no consistency imposed that requires a subsampled set of points to have the same probability density that one would find by marginalizing the surface distribution over the sample points not in the subsampling.
\end{itemize}

\subsection{Scaling consistency}

The consistency condition mentioned in the last section, which must
be imposed on probability distributions for continuous fields is:
\begin{quote}
{\em Scaling of sample points consistency:  For $ S \subset A$ indices of
discrete field variables,
\be
P(X_{S}) = \int P(X_{A}) \, dX_{A \setminus S} \label{eq:scaling}
\ee
}
\end{quote}
Note that equation~\ref{eq:scaling} is a condition which must be imposed
on the distributions which any modelling system learns where it is sensible to
supersample or subsample the field arbitrarily, as in the continuous field basis case.

\subsection{Elements of the paradigm}

The rest of this paper discusses an approach to continuous field inference which corrects the deficiencies, including the intermediate value and scaling problems, of traditional discrete-basis approaches to the inference of discrete height fields, for example.  The new approach is here named the {\it Generalized Kalman Filter}.

There are four central objects of importance within the inference approach described in this paper, one of which is a new object to Bayesian inference:
\begin{itemize}
\item
The {\bf prior} distribution for field.  The prior holds all information about
fields before any data is observed.
\item
The {\bf likelihood} distribution. The likelihood is predictive for data, given the field.  It incorporates all of the physics of the measurement process.
\item
The {\bf posterior} distribution.  The posterior distribution summarizes everything knowable about the field given assumptions of likelihood form, the prior knowledge, and all data.
\item
The {\bf knowledge-representation} (KR) distribution. Within the usual Bayesian point of view, the KR distribution is the new mathematical object.  In the paradigm described in this paper the KR distribution is the object updated when new data arrives. The KR distribution is parameterized by {\em maximally informative statistics} (see~\cite{WOLF99}) for the {\em learned} field knowledge.
Note that because the KR distribution has a finite number-of-values limitation, the KR distribution is {\em not necessarily} able to represent what {\em could have been learned} from data about the (continuous) field.  Generally, the prior distribution and the KR distribution determine an approximation (possibly exact) to the field posterior distribution.
It should be noted that modern computer architecture (memory and space-time) constraints appear to be the fundamental physical drivers for the utilization of the KR distribution, simply because storing the exact posterior generally requires an infinite amount of memory.

In the height field inference application discussed later the KR distribution is parameterized by heights at a set of discrete basis points, but holds knowledge about a continuous basis height field.  However, generally, the KR distribution may use an arbitrary set of basis functions.

One advance of the GKF is that the KR distribution is naturally adaptive in both dimension and scale, allowing the learning of continuous-basis field information at the appropriate scale, where appropriate.
\end{itemize}
Benefits of the approach described in this paper are that it has these information theoretically optimal features:  1.  A location-dependent adaptive and scalable multigrid-like algorithm, so that only the bytes necessary to represent the learned information are stored, leading to a style of maximally sparse representation of surface knowledge;  2.  A recursive updating algorithm. It will become clear that the Bayesian GKF field inference paradigm also has these properties:
\begin{itemize}
\item
It is the {\em information learned} about the field, (the KR distribution), which takes the form of a distribution over discrete values.  In the surface inference example these discrete values are heights at discrete basis points.
\item
The prior distribution for fields, in conjunction with the learned knowledge of the field held within the KR distribution determine a well-defined posterior distribution over continuous fields.
\item
The field posterior distribution is always a well defined quantity everywhere.  In the surface inference example discussed later, this continuity is at points intermediate to the discrete height field basis points of the KR distribution.
\item
The scaling condition equation~\ref{eq:scaling} is automatically imposed because the posterior distribution is a distribution over {\em fields}.
\end{itemize}
As an example consider the inference of continuous surfaces:  While it may seem obvious, in the case of continuous surface inference, that what one is actually representing with a discrete set of values in memory is only a part of the information which helps to determine the surface posterior distribution, it is unusual to {\em not} be discussing the height field as the primary representation of surface.
It is the inherently discrete nature of the storage of information in machines which forces us into this stance - {\em generally} it is impossible to represent an arbitrary {\em continuous} field with a finite set of discrete values - one must also have another object from which to compute the intermediate values of the field.  (Another way to look at the disparity between the current proposal for field inference and traditional proposals is that the traditional approaches are sufficient only for band-limited fields.)

In section~\ref{sec:paradigm} the GKF is specialized to height fields, where an example, surface representation and learning, of the GKF paradigm is described.  (The approach taken in this section is to specialize to a case that is then easily seen to generalize to the  general continuous basis field inference paradigm.)  The next section continues with observations on the update scheme.
Further sections continue with the example special case for surface distributions with particularly tractable mathematics, and final sections provide explicit forms for the general GKF equations, a discussion on their relationship to the standard Kalman filter, a discussion on the amount of information learned at each update, and a search heuristic.  Extensive appendices provide supporting mathematics for the derivations.

\section{Surface representation and inference \label{sec:paradigm}}

In this section the main ideas of the Bayesian surface representation and inference paradigm presented in this paper are given.  The technique is general, though: section~\ref{sec:observations} discusses the extension to an arbitrary-basis, arbitrary-dimension field.

\subsection{Surface distributions}

The surface and height field distributions (the prior, likelihood, and 
posterior surface and height field distributions) are discussed 
in this section.

\subsubsection{Surface and height field prior distributions} 

Consider a set $S$ of surfaces where each element $\bm{s} \in S$ is 
a height field, i.e. such that $\bm{s}=s(x,y)$ is real function
of two variables.
Write the prior probability distribution for surfaces in $S$ given the
parameters $\theta$ which
determine the prior distribution as 
\be
P(\bm{s} \mid \theta).
\ee
Consider a vector
$\bm{v} = (v_{1}, \ldots, v_{n})$ 
of discrete $(x,y)$ points, $v_{i} = (x_{i}, y_{i})$.
For any given surface $\bm{s}$
denote the associated vector of  heights by
$\bm{h}(\bm{s}, \bm{v}) = (h_{1}(\bm{s},\bm{v}), \ldots, h_{n}(\bm{s}, \bm{v}))$.
Write the prior distribution of the surface heights at the chosen points 
$\bm{v}$ as $P( {\bm{h}}_{v} \mid \theta )$.
This discrete height distribution may be found as follows:
\bea
P({\bm{h}}_{v} \mid \theta ) 
& = & \int P( {\bm{h}}_{v} \mid \bm{s}, \theta ) 
        \, P(\bm{s} \mid \theta) \, d\bm{s} 
		\label{eq:ph-v1} \\
& = & \int P( {\bm{h}}_{v} \mid \bm{s} ) \, P(\bm{s} \mid \theta) \, d\bm{s} 
		\label{eq:ph-v2} \\
& = & \int \bm{\delta}( \bm{h}_{v} - \bm{h}(\bm{s},\bm{v})) 
        \, P( \bm{s} \mid \theta) \, d\bm{s}
		 \label{eq:ph-v3}
\eea
where the vector delta-function is defined as
\bea
\bm{\delta}( \bm{h}_{v} - \bm{h}(\bm{s},\bm{v})) 
	& = & \Pi_{i=1}^{n} \delta( h_{v,i} - h_{i}(\bm{s},\bm{v}) )
\eea
Now, given that what is known is the surface heights ${\bm{h}}_{v}$
at a vector ${\bm{v}}$ of discrete $(x,y)$ points, the posterior 
distribution of surfaces is found from Bayes' theorem as
\bea
P(\bm{s} \mid {\bm{h}}_{v}, \theta) & = &
	\frac{ P({\bm{h}}_{v} \mid \bm{s}, \theta) \, P(\bm{s} \mid \theta) }
	{ P( {\bm{h}}_{v} \mid \theta ) } \label{eq:ps-hv1} \\
& = &
	\frac{ P({\bm{h}}_{v} \mid s) \, P(\bm{s} \mid \theta) }
	{ P( {\bm{h}}_{v} \mid \theta ) } \label{eq:ps-hv2} \\
& = &
	\frac{	\bm{\delta}( \bm{h}_{v} - \bm{h}(\bm{s},\bm{v})) 
		\, P(\bm{s} \mid \theta)}
	{\int \bm{\delta}( \bm{h}_{v} - \bm{h}(\bm{s},\bm{v})) 
        	\, P( \bm{s} \mid \theta) \, d\bm{s}} \label{eq:ps-hv3}
\eea
where the denominator distribution was found
in equation~\ref{eq:ph-v3}.

\subsubsection{Measurements: The Likelihood}

In general, a surface $\bm{s}$ and some other parameters $\phi$ 
not dependent upon $\bm{s}$ (i.e. camera point spread function, 
camera position and direction, lighting position and direction, 
etc.) specify the  probability distribution for data (likelihood)
\bea
P(\bm{x} \mid \bm{s}, \phi, \theta) = P(\bm{x} \mid \bm{s}, \phi) \label{eq:x-sft}
\eea
where the data distribution is  independent of $\theta$ once $\bm{s}$ is
known.  

\subsubsection{Conditioning on data: Surface and height field posterior distributions}

Given data,  the surface posterior distribution is inferred 
using Bayes' theorem as
\bea
P(\bm{s} \mid \bm{x}, \phi, \theta ) & = & 
	\frac{	P( \bm{x} \mid \bm{s}, \phi, \theta) \, P(\bm{s} \mid \phi, \theta) }
	{ P(\bm{x} \mid \phi, \theta) }
	\label{eq:surfGivenData1} \\
& = &
	\frac{	P( \bm{x} \mid \bm{s}, \phi) \, P(\bm{s} \mid \theta) }
	{ \int P( \bm{x} \mid \bm{s}, \phi) \, P(\bm{s} \mid \theta) \, d\bm{s} }
	\label{eq:surfGivenData2}
\eea
The distribution of the surface posterior marginalized to a set of discrete points
may be written using
equations~\ref{eq:surfGivenData1}--\ref{eq:surfGivenData2},
doing steps similar to those 
taken in equations~\ref{eq:ph-v1}--\ref{eq:ph-v3}, as
\bea
P( {\bm{h}}_{v} \mid \bm{x}, \phi, \theta )
& = & \int P( {\bm{h}}_{v} \mid \bm{s}, \bm{x}, \phi, \theta ) 
        \, P(\bm{s} \mid \bm{x}, \phi, \theta) \, d\bm{s} \label{eq:ph-vxpt1} \\
& = & \int P( {\bm{h}}_{v} \mid \bm{s} ) 
        \, P(\bm{s} \mid \bm{x}, \phi, \theta) \, d\bm{s} \label{eq:ph-vxpt2} \\
& = & \int \bm{\delta}( \bm{h}_{v} - \bm{h}(\bm{s},\bm{v})) 
        \, P(\bm{s} \mid \bm{x}, \phi, \theta) \, d\bm{s} \label{eq:ph-vxpt3}
\eea
In steps similar to equations~\ref{eq:ps-hv1}--\ref{eq:ps-hv3}
the surface posterior when a height field is also known is given by
\bea
P(\bm{s} \mid {\bm{h}}_{v}, \bm{x}, \phi, \theta ) 
&=&
	\frac{ P({\bm{h}}_{v}, \bm{x} \mid \bm{s}, \phi, \theta) \,
	P(\bm{s} \mid \phi, \theta) }
	{ P({\bm{h}}_{v}, \bm{x} \mid \phi, \theta) }
	\label{eq:surf-h-data-1} \\
& = &
	\frac{ 
	P({\bm{h}}_{v} \mid \bm{s}) \,
	P(\bm{x} \mid \bm{s}, \phi) \,
	P(\bm{s} \mid \theta) }
	{ P({\bm{h}}_{v}, \bm{x} \mid \phi, \theta) }
	\label{eq:surf-h-data-2} \\
& = &
	\frac{ 
	\bm{\delta}(\bm{h}_{v} - \bm{h}(\bm{s}, \bm{v})) \,
	P(\bm{x} \mid \bm{s}, \phi) \,
	P(\bm{s} \mid \theta) }
	{ \int \bm{\delta}(\bm{h}_{v} - \bm{h}(\bm{s}, \bm{v})) \,
	P(\bm{x} \mid \bm{s}, \phi) \,
	P(\bm{s} \mid \theta) \, d\bm{s} }
	\label{eq:surf-h-data-3}
\eea
where we used the facts that, given a surface, the data and the surface
heights are independent, and the surface distribution is independent
of the camera and lighting parameters $\phi$.

\subsection{ Approximating the posterior }

One motivation for approximating the surface distribution
is that generally a surface is an uncountably infinite, continuous
entity, and therefore there is little else which can be done to represent
it exactly other than to go into, literally, infinite detail (requiring an
infinite supply of memory).  It is therefore useful to have an approximation 
scheme which, although finite, captures the relevant information provided 
by data.  Another excellent reason for developing an approximation is 
mathematical tractability.  Having a representation scheme which allows 
a tractable calculation of the posterior is a huge benefit for both 
computation and communication.  Finally, it is of great interest to not 
waste computational resources while representing learned surface 
information.  The solution to the surface representation
problem presented here addresses the competition for representational
resources (memory) issue in a unique
manner.

\subsubsection{The knowledge representation distribution}

The full posterior may be written in the form
\bea
P(\bm{s} \mid \bm{x}, \phi, \theta ) & = &
	\int 	P(\bm{s} \mid {\bm{h}}_{v}, \bm{x}, \phi, \theta ) \,
		P( {\bm{h}}_{v} \mid \bm{x}, \phi, \theta ) \,
		d\bm{h}_{v} \label{eq:surf-data}
\eea
where the distributions inside the integral appear in 
equations~\ref{eq:ph-vxpt1}--\ref{eq:surf-h-data-3}.
The issue of generating 
a finite representation is not yet resolved
via equation~\ref{eq:surf-data} however, 
since storing information sufficient 
to determine the distributions
$P(\bm{s} \mid \bm{x}, \phi, \theta )$, and
$P(\bm{s} \mid {\bm{h}}_{v}, \bm{x}, \phi, \theta )$
generally requires storing an infinite set of values
in a finite amount of memory, or requires that all 
data be stored, disallowing any discarding of data and
the incremental updating of the representation.
Instead, consider the following approximation
where the prior conditioned on a set of heights, along with a new distribution, 
the {\em knowledge representation} 
distribution $\hat{P}( {\bm{h}}_{v} \mid \bm{x}, \phi, \theta )$,
are substituted for the distributions 
inside the integral of equation~\ref{eq:surf-data}.
\bea
\hat{P}(\bm{s} \mid \, \hat{P}( {\bm{h}}_{v} \mid \bm{x}, \phi, \theta ) \, )
& = &
	\int 	P(\bm{s} \mid {\bm{h}}_{v}, \theta ) \,
		\hat{P}( {\bm{h}}_{v} \mid \bm{x}, \phi, \theta ) \,
		d\bm{h}_{v} \label{eq:surf-data-approx}
\eea
It is important to note at this point that {\it any} suitable surface distribution may be substituted into the right-hand side of equation~\ref{eq:surf-data-approx} for $P(\bm{s} \mid {\bm{h}}_{v}, \theta )$, since it is important only that the resulting integral be capable of making a good approximation to the true posterior. Further, it is not necessary to restrict the basis $\bm{v}$ to discrete height field basis points, any suitable basis may be taken, for instance Fourier components. Although all of the calculations of this paper are carried thru with the form of~\ref{eq:surf-data-approx}, other forms may prove more convenient, and it is not difficult to suggest others.  In particular, since equation~\ref{eq:surf-data-approx} will be used in an iterative update loop later, updates that take for the right-hand side prior term the last posterior term appear quite reasonable (the corresponding GKF update equations may be found immediately from those presented later).

Although conditioning on the KR distribution
$\hat{P}( {\bm{h}}_{v} \mid \bm{x}, \phi, \theta )$
may seem strange, a good way to understand the meaning
is that it is the KR distribution 
which is being used as a statistic for the learned
surface information.  
The key thing to notice in equation~\ref{eq:surf-data-approx} is that, 
with reasonable regularity conditions, choosing the points of $\bm{v}$ sufficiently dense, the approximation desired to the full posterior
may become arbitrarily good. 
The trick will be to choose $\bm{v}$
appropriately, properly weighting the competing need to approximate
arbitrarily well everywhere with the limited resources that are imposed
when a finite amount of storage is available, i.e. when the dimensionality
of $\bm{v}$ is fixed.  This will be
addressed in the next section.
In the case of simple imaging systems, the point spread function and pixel
diameter are good indicators of the necessary sampling scale for $\bm{v}$.
In the super-resolved case, the resolution expected available from the data
is the appropriate scale for $\bm{v}$.

The approximation to the posterior 
of~\ref{eq:surf-data-approx} has several properties
which make it valuable:  
\begin{itemize}
\item
The prior distribution 
$P(\bm{s} \mid {\bm{h}}_{v}, \theta )$ which supplies the uncertainties
associated with points of the surface not in the vector $\bm{v}$ may be chosen to have a
simple form (see appendix~\ref{ap:construct-prior}) that is easily encoded
algorithmically in finite memory.  
\item
There is a clear separation between what was already
known - the prior $P(\bm{s} \mid {\bm{h}}_{v}, \theta )$, and what has
been learned - the KR 
distribution $\hat{P}( {\bm{h}}_{v} \mid \bm{x}, \phi, \theta )$. 
\item
There is a clear description of the scale at which information has 
been acquired in terms of the density and uncertainties associated
with the points
$(\bm{v}, h(\bm{s}, \bm{v}))$ on the surface, 
and in terms of the uncertainties of
their positions as encoded in the KR distribution.
\end{itemize}

In practice, it is useful to take a multinormal distribution over the discrete-point height field as the KR distribution.  Let the parameterization of the 
KR distribution be $\Theta_{v}$. 
For example, if the KR is taken to be 
multinormal then the parameters of that distribution are
\be
\Theta_{v}(\bm{x}) = 
(\bm{\mu}_{v}(\bm{x}),\Sigma_{v}(\bm{x})),
\ee
the mean and covariance matrix of the multinormal, 
where the functional dependence on $\bm{x}$ indicates 
a data dependency through the update procedure, and
the subscript $\bm{v}$ indicates that the parameters
parameterize a distribution of heights at points $\bm{v}$.
Because the KR distribution and its parameters
are related by a one-to-one mapping,
re-write equation~\ref{eq:surf-data-approx} as
\bea
\hat{P}(\bm{s} \mid \Theta_{v}, \theta )
& = &
	\int 	P(\bm{s} \mid {\bm{h}}_{v}, \theta ) \,
		\hat{P}( {\bm{h}}_{v} \mid \Theta_{v} ) \,
		d\bm{h}_{v}. \label{eq:mulltinfer2}
\eea
In summary, we have arrived at an approximation
to the surface posterior distribution, 
via the KR distribution, 
parameterized by $\Theta_{v}$.

\subsection{Updating the knowledge representation}

Now we discuss updating $\Theta_{v}$  when new data are acquired.  Temporarily restrict attention to the fixed $\bm{v}$ case.  During this and the next sections refer to figure~1 for a flowchart of the general GKF update process.

\subsubsection{Bayes' theorem}

Having acquired $\Theta_{v}^{n} = \Theta_{v}(x^{n})$,
from previously seen data $\bm{x}^{n}=(\bm{x}_{1}, \ldots, \bm{x}_{n})$ 
and upon seeing new data $\bm{x}_{n+1}$, the goal is to find
$\Theta_{v}^{n+1}$ such that the surface distribution
given $\Theta_{v}^{n+1}$ approximates
the surface distribution given $\bm{x}_{n+1}$
and $\Theta_{v}^{n}$.
Given new data $\bm{x}_{n+1}$
in the context of the previously seen data $\bm{x}^{n}$ summarized by
$\Theta_{v}^{n}$,
our updated surface distribution is found via Bayes' theorem
\bea
\hat{P}(\bm{s} \mid \bm{x}_{n+1}, \Theta_{v}^{n}, \phi, \theta) & = &
	\frac{ P(\bm{x}_{n+1} \mid \bm{s}, \Theta_{v}^{n}, \phi, \theta) 
		\hat{P}(\bm{s} \mid \Theta_{v}^{n}, \phi, \theta) }
{ \hat{P}(\bm{x}_{n+1} \mid \Theta_{v}^{n}, \phi, \theta) } \nn
\\
& = &
	\frac{ P(\bm{x}_{n+1} \mid \bm{s}, \phi) 
		\hat{P}(\bm{s} \mid \Theta_{v}^{n}, \theta) } 
	{ \hat{P}(\bm{x}_{n+1} \mid \Theta_{v}^{n}, \phi, \theta) } \nn
\\
& = &
	\frac{ P(\bm{x}_{n+1} \mid \bm{s}, \phi) 
		\hat{P}(\bm{s} \mid \Theta_{v}^{n}, \theta) } 
	     { \int P(\bm{x}_{n+1} \mid \bm{s}, \phi) 
		\hat{P}(\bm{s} \mid \Theta_{v}^{n}, \theta) \, d\bm{s} } \label{eq:bayes-update}
\eea
where we defined
\be
\hat{P}(\bm{x}_{n+1} \mid \Theta_{v}^{n}, \phi, \theta) =
	\int P(\bm{x}_{n+1} \mid  \bm{s}, \phi) 
		\hat{P}(\bm{s} \mid \Theta_{v}^{n}, \theta) \, d\bm{s} .
\ee
The updated posterior
$\hat{P}(\bm{s} \mid \Theta_{v}^{n}, \bm{x}_{n+1}, \phi, \theta)$
will be approximated by the $\Theta_{v}^{n+1}$ parameterized 
KR distribution of equation~\ref{eq:mulltinfer2} as
\bea
\hat{P}(\bm{s} \mid \Theta_{v}^{n+1}, \theta )
& = &
\int 	P(\bm{s} \mid {\bm{h}}_{v}, \theta ) \,
	\hat{P}( {\bm{h}}_{v} \mid \Theta_{v}^{n+1}) \,
	d\bm{h}_{v}. \label{eq:mulltinfer3}
\eea
The approximation condition for determining 
$\Theta_{v}^{n+1}$ is then written
\bea
\hat{P}(\bm{s} \mid \Theta_{v}^{n+1}, \theta )
& \approx &
\hat{P}(\bm{s} \mid \bm{x}_{n+1}, \Theta_{v}^{n}, \phi, \theta)
\label{eq:approx}
\eea
Equation~\ref{eq:approx} suggests we try to minimize various measures of the closeness of the two distributions. For example, one measure is the average square difference of the two distributions,
\be
\int \left| P_{1}(\bm{s})-P_{2}(\bm{s}) \right|^{2} d\bm{s}
\ee
but there is (apparently) no good first-principles reason to use this form.
In the next section we discuss the measure of distance which lead\bm{s} to the
{\em maximally informative} choice of $\Theta_{v}^{n+1}$.

\subsubsection{Maximally informative inference}

The measure of distance which leads to the $\Theta^{n+1}$ providing the most
information about the surface distribution is the {\em maximally informative}
choice for the statistic $\Theta^{n+1}$. 
The condition for being maximally informative, see \cite{WOLF99}, 
is that the Kullback-Leibler
distance $D( P_{1}(\bm{s}), P_{2}(\bm{s}))$ is minimized, where
\bea
D( P_{1}(\bm{s}), P_{2}(\bm{s})) = \int P_{1}(\bm{s}) \, 
log\left(\frac{P_{1}(\bm{s})}{P_{2}(\bm{s})}\right) d\bm{s}
\label{eq:kullback}
\eea
and where the $P$'s above are {\sl posterior} distributions of field, that is
\bea
P_{1}(\bm{s}) &=& \hat{P}(\bm{s} \mid \bm{x}_{n+1}, \Theta_{v}^{n}, \phi, \theta) 
\label{eq:phat_old} \\
P_{2}(\bm{s}) &=& \hat{P}(\bm{s} \mid \Theta_{v}^{n+1}, \theta ). \label{eq:phat_new}
\eea
That is,
\newline

\noindent
{\em Find the $\Theta^{n+1}$ such that}
\bea
\partial_{\Theta_{v}^{n+1}} 
\int \hat{P}(\bm{s} \mid \Theta_{v}^{n}, \bm{x}_{n+1}, \phi, \theta)  \, 
log\left(
\frac{\hat{P}(\bm{s} \mid \Theta_{v}^{n}, \bm{x}_{n+1}, \phi, \theta) }
{\hat{P}(\bm{s} \mid \Theta_{v}^{n+1}, \theta )}\right) \,
d\bm{s} = \bm{0} \nn \label{eq:kullback2} \\
\eea
{\em while at the $\Theta_{v}^{n+1}$ 
satisfying the derivative condition above}
\bea
det\left[ \partial^{2}_{\Theta_{v}^{n+1}} \int \hat{P}(\bm{s} \mid 
\Theta_{v}^{n}, \bm{x}_{n+1}, \phi, \theta)  \, 
log\left(
\frac{\hat{P}(\bm{s} \mid \Theta_{v}^{n}, \bm{x}_{n+1}, \phi, \theta) }
{\hat{P}(\bm{s} \mid \Theta_{v}^{n+1}, \theta )}\right) \,
d\bm{s} \right] < 0 \nn \label{eq:curv} \\
\eea
{\em i.e., the hessian is negative definite
and the extremum is a local maximum. 
If possible, choose the global maximum.}
Note that the Kullback-Leibler distance is asymmetric.  Generally, it is highly relevant which distribution contains the prior information and which distribution is being updated.  Maximum entropy techniques reverse the roles of $P_{1}$ and $P_{2}$ which appear here.  For a detailed explanation see~\cite{WOLF99}.

In the following section are some observations on the approach taken to maximally informative surface inference.  Section~\ref{sec:assumed-forms} then briefly makes explicit the specific distribution forms which are assumed. The Generalized Kalman Filter update equations for the surface inference example which follow from this approach are then presented in section~\ref{sec:gkf}, completing the derivation of the maximally informative approach.

\section{Observations on the update scheme \label{sec:observations}}

Note the following:
\begin{itemize}
\item
The updating scheme described here is a maximally informative update scheme
and is related to the Kalman filter.  The Kalman filter is a minimum variance
filtering scheme applicable in the case of fixed representation dimension.
The crucial step which has been taken in the current
work is the step of allowing the representation scheme to be adaptable.
We have adopted the label ``Generalized Kalman Filter'' (GKF) to describe the
idea represented here. The GKF equations are presented in section~\ref{sec:gkf}.
\item
To this point we have only optimized over $\Theta_{v}$. 
It is clear that we may also vary the number of vertices $\left| \bm{v} \right|$ of
the representation, allowing optimization over the number of vertices.
Varying the number of vertices of the representation is absolutely necessary 
if surface knowledge
at scales smaller than the current set of vertices represents
is to ever accumulate. In section~\ref{sec:gkf} the GKF update equations are derived assuming that the number of vertices in the representation basis vertex set is arbitrary at each update.
\item
Beyond allowing the number of vertices to vary, the positions
of the vertices may be allowed to vary. In section~\ref{sec:gkf} the GKF update equations are derived assuming that the representation basis vertex set positions are arbitrary.
\item
Detecting when and where new vertices are necessary is a matter of
observing directly in equations~\ref{eq:kullback} 
or~\ref{eq:kullback2} when new data 
produces a lower surface uncertainty over a region, and when 
having smaller uncertainty at neighboring vertices is not
sufficient to represent this lower uncertainty over the region.
\item 
The vertex representation for the surface knowledge is convenient,
but not necessary.  For example it is possible to extend a height field to
a height-and-reflectance field or ``arbitrary dimension field'', where the reflectance lies within a many-dimensional space.  
Reasonable structures for the covariance matrix
allow differing correlations between reflectance values and between height
values. It will be seen in in section~\ref{sec:gkf} that the GKF update equations are easily used in the ``arbitrary dimension field'' context.
\item
In its most abstract form, instead of having a ``field'',
there is simply a set of objects, while for each ``object'' 
there is an associated vector of properties,
where some of the components of the property 
vector may be considered a
location in space.  In this fairly abstracted setting, 
the collection of objects has an associated
joint probability distribution which describes the probability
distribution over configurations of objects.
It will be seen in in section~\ref{sec:gkf} that the GKF update equations are easily understood in the ``object'' context.
\item
Equation~\ref{eq:kullback2}
which defines the quantity to be minimized is where a penalty term which
indicates how many bits in hardware is available in trade for
each bit of information learned from data.  For example, one might penalize
the KL distance by $1/10$th the number of bytes it takes to represent the
new information gained by extending the number of points represented.
The exact form of the information learned about the surface distribution contained in the KR distribution is found in section~\ref{sec:learned-info}, where the dimensionality of the representation enters directly, and where bits-used penalty-terms may be introduced.
\item
The previous note points out how a minimum description length 
method fails for this problem.   It is certainly the case that that our update scheme may require much more memory (in bits) to represent the information learned than the information learned (in bits).  At some point, if information at small enough scales is desired, MDL would truncate and stop.  Clearly, applying MDL would then be a disaster.  On the other hand, what seems to work here may be called an adaptive MDL approach.
\item
Note that a method like maximum entropy is entirely
deficient for providing distributions of surfaces: given
the constraints implied by the knowledge of the distribution of the 
heights at discrete points:  maximum entropy 
ignores correlations between nearby surface
points no matter how close, an entirely ludicrous situation. On the other hand, a method like relative maximum entropy, based on inverting the roles of the distributions in equation~\ref{eq:kullback}, claims to provide the {\sl least} informative inference relative to the prior information, a heuristic, difficult to justify, at best.  Further, such approaches are typically based on likelihood distributions, rather than the posteriors that appear in equation~\ref{eq:kullback}.

\end{itemize}

\section{Surface Distribution Forms\label{sec:assumed-forms}}

\subsection{Prior}

For simplicity of mathematical presentation {\it only}, the prior in our surface inference example is taken multinormal over continuous, smooth height fields.  One particular, conveniently chosen, representation of the prior distribution is constructed in appendix~\ref{ap:construct-prior}.  This prior may be written in the
shorthand
\be
P(\bm{s} \mid \theta) = N(\bm{\mu}_{s}, \Sigma_{s})(\bm{s})
\ee
where $\theta=(\bm{\mu}_{s},\Sigma_{s})$ is the parameter vector.
The density of the height field determined by the prior
\bea
P(\bm{h}_{v} \mid \theta) &=& \int {P( \bm{h}_{v} \mid \bm{s})} \, P(\bm{s} \mid \theta) \, d\bm{s} \\
&=& \int \bm{\delta}(\bm{h}_{v}-\bm{h}(\bm{s}, \bm{v})) \, P(\bm{s} \mid \theta) \, d\bm{s} \\
&=& N(\bm{\mu}_{v}, \Sigma_{v})(\bm{h}_{v})
\eea
where
\bea
& \bm{\mu}_{v} = {A_{v s}} \bm{\mu}_{s} \nn \\
& \Sigma_{v} = {A_{v s}}  \Sigma_{s} {A_{v s}^{T}} \label{eq:project-stov}
\eea
and the projection onto the height field is given by ${A_{v s}}$. Note that equation~\ref{eq:project-stov} implies that the surface density covariance is represented differently than a discrete surface distribution covariance matrix. Specifically, the projection matrix ${A_{v s}}$ is a delta-function-like operator, and $\Sigma_{s}$ is a continuous function of two positions.  In appendix~\ref{ap:construct-prior} we show that the surface density has a compact continuous power spectrum representation, and there give the explicit form of that representation. Thus the notation of equation~\ref{eq:project-stov} must be considered a shorthand for the underlying continuous construct.

\subsection{Likelihood}

When measurement is modelled as a linear process corrupted
by gaussian noise we have
\bea
\bm{x} 	& = & M \bm{s} + \epsilon \nn \\
\epsilon & \sim & N(\bm{0},\Sigma_{\epsilon}). 
\eea
or
\bea
P( \bm{x} \mid \bm{s}, \phi) = N(M \bm{s},\Sigma_{\epsilon})(\bm{x})
\eea
where $\phi = (M, \Sigma_{\epsilon})$ is the parameter vector.

\section{The Generalized Kalman Filter equations. \label{sec:gkf}}

In this section a concise derivation of the Generalized Kalman Filter update equations specialized to the discrete basis multinormal KR distribution of equation~\ref{eq:mulltinfer2} are derived.  The updated KR need not have the same basis dimension nor position as the previous KR basis, solving the problem of how to allow updates from one representation to the next, same, finer or coarser, representation.

Proceeding, the KR distribution in terms of the parameterized height field of equation~\ref{eq:mulltinfer2} is
\be
\hat{P}(\bm{s} \mid \Theta_{v}^{n}, \theta) =
\int P(\bm{s} \mid \bm{h}_{v}, \theta) \, {\hat{P}(\bm{h}_{v} \mid \Theta_{v}^{n})} \, d\bm{h}_{v}
\ee
The distribution of surface given the height field from equation~\ref{eq:ps-hv3} is
\bea
P(\bm{s} \mid \bm{h}_{v} \theta) &=& \frac{ {P( \bm{h}_{v} \mid \bm{s})}
 \, {P(\bm{s} \mid \theta)} } {P(\bm{h}_{v} \mid \theta)} \nn \\
&=& \frac{ {\bm{\delta}(\bm{h}_{v}-\bm{h}(\bm{s}, \bm{v}))} \, {P(\bm{s} \mid \theta)} } {P(\bm{h}_{v} \mid \theta)}
\eea
Simplify the integral of the KR distribution to find
\bea
\hat{P}(\bm{s} \mid \Theta_{v}^{n}, \theta) &=&
\int \frac{ {P( \bm{h}_{v} \mid \bm{s})}
 \, {P(\bm{s} \mid \theta)} } {P(\bm{h}_{v} \mid \theta)} \, 
{\hat{P}(\bm{h}_{v} \mid \Theta_{v}^{n})} \, d\bm{h}_{v} \nn \\
&=& {P(\bm{s} \mid \theta)} \int {\bm{\delta}(\bm{h}_{v}-\bm{h}(\bm{s}, \bm{v}))} \frac{{\hat{P}(\bm{h}_{v} \mid \Theta_{v}^{n})}}{P(\bm{h}_{v} \mid \theta)} \, 
 \, d\bm{h}_{v} \nn \\
&=& {{P(\bm{s} \mid \theta)} \frac{{\hat{P}({\bm{h}(\bm{s}, \bm{v}}) \mid \Theta_{v}^{n})}}{P({\bm{h}(\bm{s}, \bm{v}}) \mid \theta)}} \label{eq:know-rep-simp}
\eea
Note how the full surface distribution is simply modified by the ratio
\bea
\frac{{\hat{P}({\bm{h}(\bm{s}, \bm{v}}) \mid
  \Theta_{v}^{n})}}{P({\bm{h}(\bm{s}, \bm{v}}) \mid \theta)}
\eea
From equation~\ref{eq:bayes-update} the Bayesian update of the KR distribution is
\bea
\hat{P}(\bm{s} \mid \bm{x}_{n+1}, \Theta_{v}^{n}, \phi, \theta) &=&
\frac{ {P(\bm{x}_{n+1} \mid \bm{s}, \phi)} \, {\hat{P}(\bm{s} \mid \Theta_{v}^{n}, \theta)} }{ {\int { {P(\bm{x}_{n+1} \mid \bm{s}, \phi)} \,{\hat{P}(\bm{s} \mid \Theta_{v}^{n}, \theta)} } \, d\bm{s} } } \nn \\
&=& \frac{ {P(\bm{x}_{n+1} \mid \bm{s}, \phi)} \, {\hat{P}(\bm{s} \mid \Theta_{v}^{n}, \theta)} }{ \hat{P}(\bm{x}_{n+1} \mid \Theta_{v}^{n}, \phi, \theta) }
\eea
Rewriting the updated distribution using equation~\ref{eq:know-rep-simp} yields
\bea
\hat{P}(\bm{s} \mid \bm{x}_{n+1}, \Theta_{v}^{n}, \phi, \theta) &\propto&
P(\bm{x}_{n+1} \mid \bm{s}, \phi) \, {P(\bm{s} \mid \theta)} \times \, \frac{{\hat{P}({\bm{h}(\bm{s}, \bm{v}}) \mid
  \Theta_{v}^{n})}}{P({\bm{h}(\bm{s}, \bm{v}}) \mid \theta)} \label{eq:simp-upd-approx} \nn \\
\eea
For maximally informative inference of the new KR we
minimize, from equation~\ref{eq:kullback},
\bea
D(P_{1}(\bm{s}), P_{2}(\bm{s})) &=& D({\hat{P}(\bm{s} \mid \bm{x}_{n+1}, \Theta_{v}^{n}, \phi, \theta)}, {\hat{P}(\bm{s} \mid \Theta_{\ov{v}}^{n+1}, \theta))} \nn \\
&=& \int {\hat{P}(\bm{s} \mid \bm{x}_{n+1}, \Theta_{v}^{n}, \phi, \theta)} \, log \left(\frac{{\hat{P}(\bm{s} \mid \bm{x}_{n+1}, \Theta_{v}^{n}, \phi, \theta)}}{{\hat{P}(\bm{s} \mid \Theta_{\ov{v}}^{n+1}, \theta)}} \right) \, d\bm{s}
 \nn \\
\eea
Note that it is not assumed here that $\bm{v}$ and $\ov{\bm{v}}$ have the same dimension.
Expanding the probability distributions within the logarithm appearing above yields
\bea
D(P_{1}(\bm{s}), P_{2}(\bm{s})) &=& 
\int {\hat{P}(\bm{s} \mid \bm{x}_{n+1}, \Theta_{v}^{n}, \phi, \theta)} \nn \\
& & \quad \times \left[ \,
 - log \left(
			{P( \bm{h}(\bm{s}, \bm{v}) \mid \theta )}
     \right)
				\right . \nn \\
& & \quad \quad \left.
 + log \left(
			{P( \bm{h}(\bm{s}, \ov{\bm{v}}) \mid \theta )}
     \right)
				\right . \nn \\
& & \quad  \quad \left.
 + log \left( 
			{P(\bm{x}_{n+1} \mid \bm{s}, \phi)}
       \right)
                 \right. \nn \\
& & \quad  \quad \left.
 - log \left( 
          	{ \hat{P}(\bm{x}_{n+1} \mid \Theta_{v}^{n}, \phi, \theta) }
        \right)
                 \right. \nn \\
& & \quad  \quad \left.
 + log \left(
			{\hat{P}( \bm{h}(\bm{s}, \bm{v}) \mid \Theta_{v}^{n} )}
       \right)
                 \right. \nn \\
& & \quad \quad  \left.
 - log \left(
			{\hat{P}( \bm{h}(\bm{s}, \ov{\bm{v}}) \mid \Theta_{\ov{v}}^{n+1} )}
       \right)
             \,  \right]
 \, d\bm{s} \label{eq:six-d-terms}
\eea
Each term has the form of an information (or uncertainty).  
Together the six terms paint a descriptive picture of how information is acquired by the maximally informative update when taken as three groups of
two terms:
Denote by ``new KR'' the two terms with $\ov{\bm{v}}$ and $\Theta_{\ov{v}}^{n+1}$, by ``previous KR'' the two terms with $\bm{v}$ and $\Theta_{v}^{n}$ and no data, and by ``new data'' the two terms with data dependency.
Now, noting the signs on these quantities, because $D$ is positive, the whole point of choosing a good $\Theta^{n+1}$ approximation by minimizing $D$ is that
\bea
& & \mbox{\em{Expected information in new KR}} \simeq \nn \\
& & \quad \quad \left(  \mbox{\it{Expected information in previous KR}} \right. \nn \\
& & \quad \quad \quad \quad \left.  
+ \mbox{\it{\it{Expected information in new data}}} \right) 
\eea
or in very rough terms we may see the update as capturing the sum-total of the available knowledge
\bea
\mbox{\it{Total knowledge}} = 
\mbox{\it{Prior knowledge}} 
+  \mbox{\it{New knowledge from data}} 
\eea
Because only terms depending upon the update parameters $\ov{\bm{v}}$ and $\Theta_{\ov{v}}^{n+1}$ are needed to perform the minimization, we drop the other terms at this point, and  after making the multinormal substitutions for the distributions in the above we have
\bea
\bar{D}(P_{1}(\bm{s}), P_{2}(\bm{s}))
&=& 
\int
 {\hat{P}(\bm{h}_{\overline{v}} \mid \bm{x}_{n+1}, \Theta_{v}^{n}, \phi, \theta)} \,
   log \left(
         { N(\bm{\mu}_{\ov{v}},\Sigma_{\ov{v}})(\bm{h}_{\ov{v}}) }
       \right)
 \, d\bm{h}_{\ov{v}} \nn \\
&-&
\int
 {\hat{P}(\bm{h}_{\overline{v}} \mid \bm{x}_{n+1}, \Theta_{v}^{n}, \phi, \theta)} \,
   log \left(
         { N(\bm{\mu}_{\ov{v}}^{n+1},\Sigma_{\ov{v}}^{n+1})(\bm{h}_{\ov{v}}) }
       \right)
 \, d\bm{h}_{\ov{v}} \nn \\ \label{eq:two-d-terms}
 \eea
To simplify the $\hat{P}$'s appearing in equation~\ref{eq:two-d-terms},  the distribution of surface given old knowledge and new data, marginalized to the height field $\ov{\bm{v}}$, is useful, as is seen by observing equations~\ref{eq:six-d-terms} and~\ref{eq:two-d-terms}.
Thus, consider 
\bea
\hat{P}(\bm{s} \mid \bm{x}_{n+1}, \Theta_{v}^{n}, \phi, \theta) &\propto&
{N( M(\phi) \bm{s}, \Sigma_{{\epsilon}}^{n+1})(\bm{x}_{n+1})} \, {N(\bm{\mu}_{s}, \Sigma_{s})(\bm{s})} \nn \\
& & \times \, \frac{N(\bm{\mu}_{v}^{n}, \Sigma_{v}^{n})(\bm{h}(\bm{s}, \bm{v}))}{N(\bm{\mu}_{v}, \Sigma_{v})(\bm{h}(\bm{s}, \bm{v}))}
\eea
found by making substitutions into~\ref{eq:simp-upd-approx} for the assumed distributions.
Since it is not necessarily the case that ${v}_{i} \in \{\ov{{v}}_{j}\}$ or that $\ov{{v}}_{i} \in \{{v}_{j}\}$. proceed by marginalizing to the union of the components of $\bm{v}$ and $\bm{\ov{v}}$, which we denote $\bm{v} \cup \bm{\ov{v}}$, and then to the $\bm{\ov{v}}$ components. Let $A_{v \cup \ov{v}, s}$ denote the projection from $\bm{v}_{s}$ to $\bm{v} \cup \bm{\ov{v}}$, $A_{\ov{v},v\cup\ov{v}}$ denote the projection from $\bm{v} \cup \bm{\ov{v}}$ to $\bm{\ov{v}}$, and $A_{\ov{v},v}$ denote the projection from $\bm{v}$ to $\bm{\ov{v}}$.
In performing the two projections (from $\bm{v}_{s}$ to $\bm{v} \cup \bm{\ov{v}}$, and then from $\bm{v} \cup \bm{\ov{v}}$ to $\bm{\ov{v}}$) in order we find (not necessarily in most simple form), using results of appendices~\ref{ap:moments}--\ref{ap:mult-mult}, that
\bea
\int \hat{P}(\bm{s} \mid \bm{x}_{n+1}, \Theta_{v}^{n}, \phi, \theta) \, d\bm{s} \setminus \ov{\bm{v}} = N(\bm{\mu}_{R}, \Sigma_{R})(\bm{h}_{\ov{\bm{v}}})
\eea
where
\bea
&\bm{\mu}_{\ov{v}}^{R} = 
  \Sigma_{R} 
  ( 
     \Sigma_{Q}^{-1}
       \bm{\mu_{\ov{v}}}^{Q}
   + (\Sigma_{\ov{v}}^{n})^{-1}
       \bm{\mu_{\ov{v}}^{n}}
   - \Sigma_{\ov{v}}^{-1}
       \bm{\mu_{\ov{v}}}
 ) \nn \\
&\Sigma_{R}^{-1} =
     \Sigma_{Q}^{-1}
   + (\Sigma_{\ov{v}}^{n})^{-1}
   - \Sigma_{\ov{v}}^{-1} \label{eq:general-kalman}
\eea
and where
\bea
&\bm{\mu}_{\ov{v}}^{Q} = A_{\ov{v},v\cup\ov{v}} A_{v\cup\ov{v},s} \bm{\mu}_{s}^{P}
  \nn \\
& \Sigma_{Q}^{-1} = 
    A_{\ov{v},v\cup\ov{v}} A_{v\cup\ov{v},s}
      \Sigma_{P}^{-1}
    A_{v\cup\ov{v},s}^{T} A_{\ov{v},v\cup\ov{v}}^{T} \nn \\ \\
&\bm{\mu}_{s}^{P} = 
  \Sigma_{P} 
  (
    \Sigma_{s}^{-1} \bm{\mu}_{s} + M^{T} \Sigma_{\epsilon}^{-1} \bm{x}_{n+1}
  ) \nn \\
& \Sigma_{P}^{-1} =
  \Sigma_{s}^{-1} + M^{T} \Sigma_{\epsilon}^{-1} M \label{eq:kf} \nn \\ \\
& \bm{\mu_{\ov{v}}^{n}} = A_{\ov{v},v} \bm{\mu_{v}^{n}} \nn \\
& (\Sigma_{\ov{v}}^{n})^{-1} = A_{\ov{v},v} (\Sigma_{v}^{n})^{-1}
  A_{\ov{v},v}^{T} \nn \\ \\
& \bm{\mu_{\ov{v}}} = A_{\ov{v},v} \bm{\mu_{v}} \nn \\
& \Sigma_{\ov{v}}^{-1}=A_{\ov{v},v} \Sigma_{v}^{-1} A_{\ov{v},v}^{T} \nn \\ \\
& \bm{\mu}_{v} = A_{v,s} \bm{\mu_{s}} \nn \\
& \Sigma_{{v}}^{-1} = A_{v,s} \Sigma_{s}^{-1} A_{v,s}^{T} \nn \\
\eea
Using the results of appendix~\ref{ap:log-average}, the quantities of equation~\ref{eq:general-kalman} above correspond to the values of the mean and standard deviation parameters of the new KR, found at the minimum Kullback Leibler distance, i.e. the minimization is immediately apparent from those results.  Thus:
\bea
& \Theta_{\ov{v}}^{n+1}=(\bm{\mu}_{\ov{v}}^{n+1}, \Sigma_{\ov{v}}^{n+1}) \nn \\
&  \bm{\mu}_{\ov{v}}^{n+1} = \bm{\mu}_{\ov{v}}^{R} \nn \\
&  \Sigma_{\ov{v}}^{n+1} = \Sigma_{\ov{v}}^{R}
\eea
Equations~\ref{eq:general-kalman} are the Generalized Kalman Filter (GKF) update equations for the surface inference example, yet are quite a bit more general (the necessary change of variables needed when the forward projection is nonlinear appears in appendix~\ref{ap:nonlinear}). Having these update equations allows one to consider updating a representation of any dimension relative to the original representation.  Thus. knowledge may be represented in finer detail, corresponding to the old representation being contained in the new, knowledge may be represented in the same detail, corresponding to the case when the new representation is the same as the old representation, or knowledge may be tossed, corresponding to the case when the new representation does not contain the old representation.  The maximally informative inference approach and its result of the Kullback Leibler distance on conditional posteriors led directly here to deriving the GKF and the solution of the problem of storing knowledge at scales adaptive to the actual needs of the data driving the update.  The standard KF is discussed in~\cite{BROW83}.

\section{Specializing the GKF}

When the surface of interest is itself a discrete height field, and the KR representation basis never changes in dimension nor position from that height field's basis, then all projections appearing in equations~\ref{eq:general-kalman} and following are identities, and the update equations simplify to the standard Kalman filter equations, in effect equations~\ref{eq:kf} only, given suitable identification of the variables.

\section{Information learned \label{sec:learned-info}}

Once a new set of parameters has been chosen, and for the purpose of evaluating the new update in the context of other possible updates at different scales, using different representational bases, it is useful to have the quantity of information about the surface distribution that is contained in the KR at the maximally informative update. Using the results of appendix~\ref{ap:log-average} in equation~\ref{eq:two-d-terms} we have this information, up to a constant, is given by
\bea
I_{R} & = & C(\bm{x}_{n+1}, \Theta_{v}^{n}, \phi, \theta) \nn \\
& + & \frac{1}{2} \left( Tr\left[ (\Sigma_{R}+U(\bm{\mu}_{R}-\bm{\mu}_{\ov{v}})) \otimes \Sigma_{\ov{v}}^{-1}\right]  + log(\left| \Sigma_{\ov{v}} \right|) \right) \nn \\
& - &  \frac{1}{2} \left( Tr\left[ \Sigma_{R} \otimes \Sigma_{R}^{-1} \right]  +  log(\left| \Sigma_{R} \right|) \right)
\eea
Note that the $d$'s (representation basis dimensions) from the
$d log(2 \pi)$'s of equation~\ref{eq:exp-info} have cancelled. However the $d$'s remain hidden within the terms as matrix dimensions. When considering optimizing learned inormation against storage resources, one must weigh a separate cost in bits for the memory used against the bits learned, the expression above. Note also, interestingly the expression above contains a BIC-like $log(d)$ dependence term.

\section{Search for update parameters}

Now that we know what the update equations for the updating of the KR distribution look like, it is worthwhile considering how an updating scheme might be implemented to acquire information at the appropriate scale.  First, we dismiss the notion that we will ever be using the continuous height field $\bm{v}_{s}$ (the support of $\bm{s}$) at any time.  None of the update equations force that to happen!  Second, since we have concluded that computationally $\bm{v}_{s}$ is a discrete set, and since there will always be pathological cases where the surface is much rougher than we care to represent, we acknowledge that fact and proceed by presenting a useful algorithm which allows the updating of the KR while maintaining the ability to explore a large range of scales.  The following multigrid-style algorithm provides the general flavor:
\begin{itemize}
\item
Choose $\bm{v}_{s}$ denser by several orders of scale than the current representation, and using other criteria associated with the knowledge of the data acquisition system (see below).
\item
Choose $\ov{\bm{v}}$ at regular scales intermediate between $\bm{v}_{s}$ and the old KR on $\bm{v}$, compute the updates on all $\ov{\bm{v}}$ chosen at these scales.
\item
Compute the information learned at each scale.
\item 
Plot the information learned as a function of increasing density (decreasing
scale).
\item
Choose, based on exploration of the plot, and costs associated with storing the learned information, whether to explore other octaves of scale.  If Choose to explore, repeat above procedure.
\item
If choice is to pick an informationally and storage attractive KR, do this and update the representation accordingly.
\end{itemize}
In the surface reconstruction problem data often comes in the form of images.  The images may come from devices with vastly different resolutions, and the known parameters of pixel size, point spread function and geometry determine the appropriate reconstruction scale.  Finally adapting the surface to resolve at sub-pixel scales requires a memory-aggressive approach which extends the exploration farther out on the learning curve towards smaller, denser representation scales.

\section{Conclusion}

Field inference has been generalized from the typical discrete fixed-basis setting to a continuous-basis setting.  The problem of surface inference was solved in the context of continuous field inference. Using the approach of acquiring the maximally informative KR distribution, the GKF equations were found.  The GKF allows the updated KR parameters to be found at any scale and/or ``positions'' (abstractly, basis components). The approach allows the learning of information at the relevant scales desired.  It provides an information-theoretic justification for location-dependent adaptive multi-grid inference.  It also effectively provides similar justification for a scale-adaptive MDL method.  This is apparently the first time that the maximally informative inference of continuous-basis objects and the multigrid approach have been rigorously justified.

\section{Acknowledgements}

I thank the members of the Ames Data Understanding group for their interest and comments, especially the invaluable valiant contributions of Dr. Robin D. Morris, who thoughtfully, carefully, and painstakingly spent a week-in-agony checking the maths (any remaining mistakes are fully mine, however), and Drs. Vadim Smelyanskiy and David Maluf for their comments.  Finally, immense thanks go to Dr. Peter Cheeseman, for comments, and support.  This project was partially supported by the NASA Ames Center for Excellence in Information Technology contract NAS-214217.

\section{Appendices}

\subsection{Construction of a 2D surface prior \label{ap:construct-prior}}

In this appendix we first introduce the reader to the fourier representation of a gaussian process, then using the notions developed find the representation for a 2D gaussian process over the plane, where the correlations of the process at points $\bm{x}$ and $\bm{y}$ are proportional to $exp(- k \left| \bm{x} - \bm{y} \right|)$, $k>0$, a simple translation-invariant choice for the form of the correlation structure of the probability density of surfaces having the plane as support. The utility for the GKF of having this process is that it serves as a simply computed algorithmic representation of the prior for surfaces having the plane as support.

\subsubsection{The discrete gaussian process}

Consider $f(n, \bm{c})$, $n \in Z_{N} = \{ -N, \ldots, -1, 0, 1, \ldots, N \}$, 
a discrete process with expression as the fourier expansion
\be
f(n,\bm{c}) = \sum_{k=-N}^{N} c_k e^{ i k n }
\label{eq:fourier-process}
\ee
where the coefficients $\bm{c}=(c_{k})$ 
are constrained by $f \in \bm{R}$ so that
$c_{k} = c_{-k}^{*}$, and the $n$ and $k$ range over $Z_{N}$.  
Let the coefficients be random variables:
$c_{k} = x_{k} + i y_{k}$ with $x_{k} \sim N(0,\sigma_{k})$ and 
$y_{k} \sim N(0,\sigma_{k})$ both gaussian distributed random variables
with mean $0$ and standard deviation $\sigma_{k}$.
Now, dropping the $k$'s, the joint density of $(x,y)$ is given by
\be
P_{x,y}(x,y) = \frac{e^{-x^2/2\sigma^{2}}}{\sqrt{2 \pi} \sigma} 
         \frac{e^{-y^2/2\sigma^{2}}}{\sqrt{2 \pi} \sigma}.
\ee
From this the joint density of $(r,\theta)$ where $r=\sqrt{x^2+y^2}$ and
$\theta = \arctan(y/x)$ is given by
\be
P_{r,\theta}(r,\theta) = \frac{r e^{-r^2/2 \sigma^2}}{2 \pi \sigma^2}.
\ee
The density of $r$ is given directly by integrating over $\theta$
\be
P_{r}(r) = \frac{r e^{-r^2/2 \sigma^2}}{\sigma^2},
\ee
while the density of $\theta$ is given directly by integrating over $r$
\be
P_{\theta}(\theta) = \frac{1}{2 \pi}.
\ee
Making a change of variables, the density of $c c^{*} = x^{2} + y^2 = r^2$
is given by the exponential distribution
\be
P_{c c^{*}}(u) = \frac{e^{-u/2 \sigma^2}}{2 \sigma^2} \label{eq:cc*}
\ee
The distribution of $c_{k} + c_{-k} = 2 Re[c_{k}] = 2 x_{k}$, $k>0$ 
is of interest because the process is real.
\be
P_{c+c^{*}}(u) = \frac{e^{-u^2/2 (2 \sigma)^{2}}}{\sqrt{2 \pi} 2 \sigma}
 \label{eq:c+c*}
\ee
which is just a gaussian with zero mean but twice the variance of the
components $x$ and $y$ of $c$.
Note that the actual coefficients in equation~\ref{eq:fourier-process}
$c_{k} e^{i k n} + c_{-k} e^{-i k n} = 2 Re[ c_{k} e^{i k n}]$ also
have the distribution of equation~\ref{eq:c+c*} since the phase of 
$c_{k}$ is uniformly 
distributed in $[ 0 , 2\pi ]$.

Now, given a set of integers $\zeta \subset Z_{N}$ we may 
ask for the density of the sampled values of the process $f$ at $\bm{\zeta}=(n_{1}, n_{2}, \ldots, n_{m})$
\be
\bm{f}(\bm{\zeta})= (f(n_{1}), f(n_{2}), \ldots, f(n_{m})), 
\ee where
$m = \left| \zeta \right|, 
n_{i} \in Z_{N}, i=1,\ldots,m$.
Define
\be
\bm{f}(\bm{\zeta}, \bm{c}) = (f(n_{1}, \bm{c}), f(n_{2}, \bm{c}), \ldots, f(n_{m}, \bm{c}))
\ee
Then the probability density function which describes the sampled values is 
\be
P(\bm{f}(\bm{\zeta})) =
\int  \delta( \bm{f}(\bm{\zeta}) - \bm{f}(\bm{\zeta}, \bm{c})) \, P(\bm{c}) \, d\bm{c}
\ee
where
\be
P(\bm{c}) = P(c_{0}) \prod_{k=1}^{N} \, P(c_{k} + c_{-k})
\ee
Note that
that the density of $P(\bm{f}(\bm{\zeta}))$ is multivariate gaussian since 
the representation of $\bm{f}(\bm{\zeta},\bm{c})$
as a fourier series shows that it is the sum of gaussian random vectors with components $2 Re[ c_{k} e^{i k n}]$. The covariances of the process are found as
\bea
\Sigma_{m,n} = E[ f(m) f(n) ] & = & E[ f(m) f^{*}(n) ] \nn \\
&=& E\left[ \sum_{k, l = -N}^{N} c_{k} c_{l}^{*} e^{i (k m - l n) } \right] \nn \\
&=&  \sum_{k= -N}^{N} E[ c_{k} c_{k}^{*}]  e^{i k (m - n) }  \nn \\
&=&  F [ E[c_{k} c_{k}^{*}] ] (m-n) 
\eea
where we used the fact that the coefficients of different frequency are uncorrelated for $k \ne l$, i.e $E[c_{k} c_{l}^{*}]=0$ for $k \ne l$. Define
the power spectrum $R(k)$ as
\be
R(k) = E[ c_{k} c_{k}^{*} ]
\ee
Then we have that the covariance is given by the fourier transform of the power spectrum,
\be
\Sigma_{m,n} = E[ f(m) f(n) ] = F[R](m-n) = \Sigma_{m-n}
\ee
where we have acknowledged that the covariance structure is dependent only upon the difference $m-n$.
From this we see that the inverse fourier transform of the covariance is the power spectrum,
\be
F^{-1}\left[ \Sigma_{u} \right](k) = R(k)
\ee
Finally, note that the density of $c_{k} c_{k}^{*}$ given by equation~\ref{eq:cc*} allows us to infer the parameters $\sigma_{k}$ which are the standard deviations of the gaussian processes $x_{k}$ and $y_{k}$ underlying the coefficients $c_{k}$, since from equation~\ref{eq:cc*}
\be
E[ c_{k} c_{k}^{*} ] = \int u \frac{e^{-u/2 \sigma_{k}^2}}{2 \sigma_{k}^2} \, du = {2 \sigma_{k}^2}
\ee
In the next section the basis for gaussian processes developed here is extended to the continuous 2D case to compute the power spectrum of a process specified by a continuous-basis covariance structure. 

\subsubsection{The continuous-basis 2D process \label{ap:cont-basis-prior}}

Similar to the development in the last section, in two dimensions, given the continuous-basis covariance $\Sigma_{\bm{x}} = exp(- k \left| \bm{x} \right|)$, $k>0$., the power spectrum is found as the inverse fourier transform of the covariance, i.e.
\bea
R(\bm{u}=(u,v)) &=& F_{2}^{-1}[ \Sigma_{\bm{x}} ](u,v) \nn \\
&=& \int \int e^{- k \left| (x,y) \right|} e^{-i u x} e^{- i v y} \, dx \, dy
\label{eq:pw-spectrum-xform}\eea
Make the change of variables $(x,y) \rightarrow (r,\theta)$ so that $x = r cos(\theta)$, $y = r sin(\theta)$, then
\be
R(u,v) =
\int_{0}^{\infty} \int_{0}^{2 \pi} e^{- k r} e^{-i r (u cos(\theta) + v sin(\theta))} \, r \, dr \, d\theta
\ee
For simplicity, make the further change of variables $(u,v) \rightarrow (s,\phi)$ so that $u = s cos(\phi)$, $v = s sin(\phi)$, so that
\bea
R(s, \phi) &=&
\int_{0}^{\infty} \int_{0}^{2 \pi} e^{- k r} e^{-i r s (cos(\phi) cos(\theta) + sin(\phi) sin(\theta))} \, r \, dr \, d\theta \nn \\
&=&
\int_{0}^{\infty} \int_{0}^{2 \pi} e^{- k r} e^{-i r s cos(\theta-\phi) } \, r \, dr \, d\theta \nn \\
&=& 
\int_{0}^{\infty} r e^{- k r} \int_{0}^{2 \pi} e^{-i r s cos(\theta-\phi) } \, d\theta \, dr \nn \\
R(s) &=& 
2 \pi \int_{0}^{\infty} r e^{- k r} J_{0}(r s) \, dr
\eea
Finally,
\bea
R(\bm{u}) &=&  \frac{2 \pi k}{(\left| \bm{u} \right|^{2}+k^{2})^{3/2}}
\eea
Note that we have neglected the proportionality constant $1/2 \pi$ in the fourier transform, amounting to normalizing the delta function to $2 \pi$, and have scaled $\bm{u}$ to units of cycles per $2 \pi$. Note also that both the covariance of the process and the power spectrum scale with the same proportionality constant.  Harmonic analysis is discussed in~\cite{ROSE74}

\subsection{Multinormal density MGF \label{ap:moments}}

The moment generating function for a probability distribution $f$ is defined
as the functional
\be
M[f](\bm{\lambda}) = E_{f}[ e^{ Tr[U(\bm{\lambda}, \bm{x})] } ]
\ee
where $U(\bm{y},\bm{z})$ is defined such that $U=[U_{ij}]$ and $U_{ij}(\bm{y},\bm{z}) := y_{i} z_{j}$, from which holds the property
\be
\frac{\partial^{k} M[f](\bm{\lambda})}{ \partial \lambda_{i_{1}} \ldots \lambda_{i_{k}} } \mid_{ \bm{\lambda}=0 } = E_{f}[ x_{i_{1}} \ldots x_{i_{k}} ]
\ee
i.e the moments are found as derivatives of the MGF with respect to the
parameter $\bm{\lambda}$ at $\bm{\lambda}=0$.

Take the multinormal density function for $\bm{x}$
\bea
P(\bm{x} \mid \Theta) &=& N(\Theta)(\bm{x}) \nn \\
&=& N(\bm{\mu}, \Sigma)(\bm{x}) \nn \\
&=&  \frac{1}{ (2\pi)^{d/2} \mid \Sigma \mid^{1/2} } 
exp( -\frac{1}{2} Tr[ U(\bm{x}-\bm{\mu}) \otimes \Sigma^{-1}  ] )
\eea
where $U(\bm{y})$ is defined such that 
$U_{ij}(\bm{y}) := U_{ij}(\bm{y},\bm{y})$ 
and $d = Dim(\bm{x})$.
The MGF of $N(\Theta)(\bm{x})$ is then given by
\bea
& & M[ N(\Theta)(\bm{x}) ](\bm{\lambda}) = E[ e^{Tr[U(\bm{\lambda}, \bm{x})]} \mid \Theta ] \nn \\
& & \quad \quad = \int 
\frac{1}{ (2\pi)^{d/2} \mid \Sigma \mid^{1/2} } exp( -\frac{1}{2} Tr[ U(\bm{x}-\bm{\mu}) \otimes \Sigma^{-1}  ] + Tr[U(\bm{\lambda},\bm{x})]) \, d\bm{x}
 \nn \\
\eea
Minus twice the exponent of  the integral above may be written as
\bea
 Tr[ U(\bm{x}-\bm{\mu}) \otimes \Sigma^{-1} ] - 2 Tr[U(\bm{\lambda}, \bm{x})]
&=& Tr[ U(\bm{x}-(\bm{\mu}-\bm{\lambda} \, \Sigma)) \otimes \Sigma^{-1} ] \nn \\
& & \, \, + Tr[ U(\bm{\mu})  \otimes  \Sigma^{-1} ] \nn \\
& & \, \, - Tr[ U(\bm{\mu}-\bm{\lambda} \, \Sigma) \otimes \Sigma^{-1} ] \nn \\
&=& Tr[ U(\bm{x}-(\bm{\mu}-\bm{\lambda} \, \Sigma)) \otimes \Sigma^{-1} ] \nn \\
& & \, \, - Tr[ U(\bm{\lambda}) \otimes \Sigma ] \nn \\
& & \, \, - 2 Tr[ U(\bm{\lambda}, \bm{\mu}) ]
\eea
from which the moment generating function is immediately found as
\be
M[ N(\Theta)(\bm{x}) ](\bm{\lambda}) = 
exp( \, Tr[ U(\bm{\mu},\bm{\lambda}) ] + \frac{1}{2} Tr[ U(\bm{\lambda}) \otimes \Sigma ] \, ) \label{eq:x-mgf}
\ee
From the above we have
\bea
& E[ x_{i} \mid \Theta] = \mu_{i} \nn \\
& E[ (x_{i}-\mu_{i}) (x_{j}-\mu_{j}) \mid \Theta] = \Sigma_{ij}
\eea
which agrees with the calculation of appendix~\ref{ap:moments}.  Two things to note: 1. The inverse of $\Sigma$ is assumed to exist. 2. All moments are determined by simple products and sums of the parameters $(\bm{\mu},\Sigma)$.

\subsection{Multinormal linear change of variables\label{sec:linear-change}}

Letting $\bm{y} = A \bm{x}$ be the change of variables, where $P(\bm{x} \mid \Theta) = N(\Theta)(\bm{x})$, the MGF of the density $P(\bm{y} \mid \Theta)$ is found from the MGF of the density for $P(\bm{x} \mid \Theta)$ in a straightforward manner as
\bea
M[ P(\bm{y} \mid \Theta) ](\bm{\lambda}) 
& = & E[ e^{Tr[U(\bm{\lambda}, \bm{y})]} \mid \Theta ] \nn \\
& = & E[ e^{Tr[U(\bm{\lambda}, A \bm{x})]} \mid \Theta ] \label{eq:changed} \\
& = & E[ e^{Tr[U(A^{T} \bm{\lambda}, \bm{x})]} \mid \Theta ] \nn \\
& = & exp( \, Tr[ U(\bm{\mu}, A^{T} \bm{\lambda}) ] + \frac{1}{2} Tr[ U(A^{T} \bm{\lambda}) \otimes \Sigma ] ) \nn \\
& = & exp( \, Tr[ U(A \bm{\mu}, \bm{\lambda}) ] + \frac{1}{2} Tr[ U(\bm{\lambda}) \otimes (A \Sigma A^{T})] ) \nn \\
\eea
Note that the dropped subscripts $x$ and $x$ of the $\Theta$ and $\lambda$ are easily determined by the context, and that the density used to take the expectation naturally changed in equation~\ref{eq:changed} from $P(\bm{y} \mid \Theta)$ to $P(\bm{x} \mid \Theta)$ without confusion. With this result and referring to equation~\ref{eq:x-mgf} and preceding we find that the density for $\bm{y}$ 
is multinormal with
\bea
& \bm{\mu_{y}} = A \bm{\mu_{x}} \nn \\
& \Sigma_{y} = A \Sigma_{x} A^{T}
\eea
Note that everywhere the condition of $A$ was neither mentioned nor assumed,  thus $A$ may be a rectangular matrix or otherwise not of full rank.

\subsection{Multinormal projections}

Another useful operation is that of projection onto a subset of the components
of the argument of the multinormal distribution.  Projections may be trivially represented as a linear operation, where the ``projection matrix'' is typically a rectangular matrix having the form of a unique (single) element of value $1$ in each row and column, zeroes elsewhere.  Finding the distribution of the projected variables is equivalent to the operation of marginalizing over the components not in the projection.  Let $A$ be the projection matrix selecting a subset of the variables of $\bm{x}$ as $\bm{y}=A \bm{x}$. Then, using the result of section~\ref{sec:linear-change}, we immediately find integrals of the form
\be
\int N(\bm{\mu}, \Sigma)(\bm{x}) \, d\bm{x}\setminus\bm{y} = N(A \bm{\mu} , A \Sigma A^{T} )(\bm{y})
\ee
Both vector $A \bm{\mu}$ and the matrix $A \Sigma A^{T}$ are now
just appropriately rearranged pieces of the original vector $\bm{\mu}$ and 
matrix $\Sigma$. Specifically, if $y_{k}=x_{i_{k}}$ then 
$[A \Sigma A^{T}]_{pq} = \Sigma_{i_{p} j_{q}}$.

\subsection{Multinormal  multiplication \label{ap:mult-mult}}

One operation which frequently occurs in Bayesian inference is that of taking the product of two multinormal distributions of the same variable and normalizing that product to find a new distribution. Finding the new $\Theta = (\bm{\mu},\Sigma)$ amounts to completing the square, but it
is useful to state the result, and we do this here.  Let $\Theta_{1} = (\bm{\mu}_{1},\Sigma_{1})$ and $\Theta_{1} = (\bm{\mu}_{1},\Sigma_{1})$ be the parameters of the multinormal distributions in the product. Then
\bea
& \bm{\mu} = \Sigma (\Sigma_{1}^{-1} \bm{\mu_{1}} + \Sigma_{2}^{-1} \bm{\mu_{2}}) \nn \\
& \Sigma = (\Sigma_{1}^{-1} + \Sigma_{1}^{-1})^{-1}
\eea

\subsection{Expected uncertainty in multinormals \label{ap:log-average}}

It is useful to know the expected uncertainty of one gaussian distribution in the context of another. Consider the quantity
\bea
E[ - log(P(\Theta_{2})(\bm{x})) \mid \Theta_{1}] = 
- \int N(\bm{\mu}_{1}, \Sigma_{1})(\bm{x}) \, log \left( N(\bm{\mu}_{2}, \Sigma_{2})(\bm{x}) \right) \, d\bm{x}
\eea
which occurs in similar form in the development of the Generalized Kalman Filter (section~\ref{sec:gkf}) and represents the expected uncertainty, or entropy, of the surface representation in the context of the updated surface distribution. The value of this integral is found straightforwardly using the results mentioned in appendix~\ref{ap:moments} as
\bea
E[ - log(N(\bm{\mu}_{2}, \Sigma_{2})(\bm{x})) \mid \Theta_{1}] 
& = &
\frac{1}{2} E\left[ Tr[ U(\bm{x}-\bm{\mu}_2) \otimes \Sigma_{2}^{-1}] \right] \nn \\
& & \quad + \frac{d}{2} log(2 \pi) + \frac{1}{2} log(\left| \Sigma_{2} \right|) \nn \\
& = & \frac{1}{2} Tr\left[ (\Sigma_{1}+U(\bm{\mu}_{1}-\bm{\mu}_{2})) \otimes \Sigma_{2}^{-1}\right] \nn \\
& & \quad + \frac{d}{2} log(2 \pi) + \frac{1}{2} log(\left| \Sigma_{2} \right|) \nn \\ \label{eq:exp-info}
\eea

\subsection{Maximizing the expected information \label{ap:log-average-max}}

Varying $\Sigma_{2}$, the minimum value of the uncertainty above occurs when $\Theta_{2} = \Theta_{1}$.  That this is true for the $\bm{\mu}$ component of $\Theta_{2}$
is immediate from the positive definite quadratic nature of the first term. For the $\Sigma$ component the following fact following from the properties of determinants and matrix inverses facilitates the result:
\be
\frac{\partial \left| \Sigma \right|}{\partial \Sigma_{kl}} = (-1)^{k+l} \frac{Cof_{kl}(\Sigma)}{\left| \Sigma \right|} = \Sigma_{kl}^{-1}
\ee

\subsection{Notes on matrix inverses and submatrices}

Given the invertible matrix $V$, composed in the following manner of submatrices $V_{11}$, $V_{12}$, $V_{21}$, $V_{22}$,
\be
A = \left[
      \begin{array}{cc}
		V_{11} & V_{12} \\
        V_{21} & V_{22}
      \end{array}
    \right]
\ee
and its inverse
\be
A^{-1} = \left[
      \begin{array}{cc}
		\hat{V}_{11} & \hat{V}_{12} \\
        \hat{V}_{21} & \hat{V}_{22}
      \end{array}
    \right]
\ee
then it is immediate that the following relationships hold among the submatrices
\bea
\left[
      \begin{array}{cc}
		I_{11} & N_{12} \\
        N_{21} & I_{22}
      \end{array}
    \right]
&=& \left[
      \begin{array}{cc}
		V_{11} \hat{V}_{11} + V_{12} \hat{V}_{21} 
      & V_{11} \hat{V}_{12} + V_{12} \hat{V}_{22} \\
        V_{21} \hat{V}_{11} + V_{22} \hat{V}_{21}
      & V_{21} \hat{V}_{12} + V_{22} \hat{V}_{22}
      \end{array}
    \right] \label{eq:inv-sub}
\eea
where $I$ and $N$ represent the identity and zero matrices respectively. Any quadratic operator $\bm{x}^{T} Q \bm{x}$ may be decomposed using projection matrices $A$ and $\ov{A}$ where these are diagonal matrices with one and zero entries only, and where
\be
A+\ov{A}=I
\ee
in the following manner
\bea
\bm{x}^{T} Q \bm{x} &=& \bm{x}^{T} (A+\ov{A}) Q (A+\ov{A})^{T} \bm{x} \nn \\
&=& \bm{x}_{A}^{T} Q_{AA} \bm{x}_{A} + \bm{x}_{A}^{T} Q_{A\ov{A}} \bm{x}_{\ov{A}} + \bm{x}_{\ov{A}}^{T} Q_{\ov{A}A} \bm{x}_{A} + \bm{x}_{\ov{A}}^{T} Q_{\ov{A}\ov{A}} \bm{x}_{\ov{A}}
\eea
Now, assume $Q$ is symmetric and that both it and $Q_{AA}$ and $Q_{\ov{A}\ov{A}}$ are invertible, and rewrite this form as the sum of two terms as follows
\bea
\bm{x}^{T} Q \bm{x} &=& (\bm{x}_{A}-\bm{\alpha})^{T} Q_{AA} (\bm{x}_{A}-\bm{\alpha}) + C(\bm{x}_{\ov{A}}) \nn \\
&=& \bm{x}_{A}^{T} Q_{AA} \bm{x}_{A} - \bm{x}_{A}^{T} Q_{A \ov{A}} \bm{x}_{\ov{A}} - \bm{x}_{\ov{A}}^{T} Q_{\ov{A} A} \bm{x}_{A} + \bm{\alpha}^{T} Q_{AA} \bm{\alpha} + C(\bm{x}_{\ov{A}}) \nn \\
\eea
where $\bm{\alpha} = (Q_{AA})^{-1} Q_{A \ov{A}} \bm{x}_{\ov{A}}$. Thus
\be
C(\bm{x}_{\ov{A}}) = \bm{x}_{\ov{A}}^{T} \left( Q_{\ov{A}\ov{A}} - Q_{\ov{A}A} (Q_{AA})^{-1} Q_{A\ov{A}} \right)  \bm{x}_{\ov{A}}
\ee
Applying the identities of equation~\ref{eq:inv-sub}
\be
Q_{AA} \hat{Q}_{A \ov{A}} + Q_{A \ov{A}} \hat{Q}_{\ov{A}\ov{A}} = N_{A \ov{A}} \nn \\
\ee
followed by
\be
Q_{\ov{A} A} \hat{Q}_{A \ov{A}} + Q_{\ov{A}\ov{A}} \hat{Q}_{\ov{A}\ov{A}} = I_{\ov{A}\ov{A}} \nn \\
\ee
find that
\be
Q_{\ov{A}\ov{A}} - Q_{\ov{A}A} (Q_{AA})^{-1} Q_{A\ov{A}} = (\hat{Q}_{\ov{A}\ov{A}})^{-1}
\ee
so that
\be
C(\bm{x}_{\ov{A}}) = \bm{x}_{\ov{A}}^{T} (\hat{Q}_{\ov{A}\ov{A}})^{-1} \bm{x}_{\ov{A}}
\ee
which immediately provides an alternate method for marginalizing gaussian distributions.

\subsection{Alternate inverse forms}

In the GKF update equations expressions for updating inverse matrices in terms of the sum of other inverse matrices occur.  Because one of the summand matrices may not be well-conditioned, it is of interest to find an expression for the updated matrix in terms of the other matrices, which explicitly is not a function of the inverse matrices. Thus, let $P$, $Q$, $R$ be invertible matrices such that
\bea
P^{-1} = Q^{-1} + R^{-1}
\eea
Then we find
\bea
P & = & Q - Q ( Q + R )^{-1} Q
\eea
by the following direct substitution
\bea
P P^{-1} & = & (Q - Q ( Q + R )^{-1} Q)(Q^{-1} + R^{-1}) \nn \\
         & = & I - Q \left[ ( Q + R )^{-1} (I + Q R^{-1}) - R^{-1} \right] \nn \\
& = & I
\eea

\subsection{Nonlinear forward projection \label{ap:nonlinear}}

In the nonlinear forward projection case the projection is given by $\bm{f}(\bm{s})$, where $\bm{f}(\cdot)$ is a nonlinear function of $\bm{s}$ rather then the linear form $M\bm{s}$. Because the
derivative of the forward projection is often a straightforward object to
compute, expand $\bm{f}(\bm{s})$ about the mean of the old surface, $\bm{\mu}_{s}$
\be
\bm{x} = \bm{f}(\bm{\mu}_{s}) + \frac{\partial \bm{f}}{\partial \bm{s}} \mid_{\bm{\mu}_{s}} ( \bm{s} - \bm{\mu}_{s}) + \epsilon
\ee
Letting $M=\frac{\partial \bm{f}}{\partial \bm{s}} \mid_{\bm{\mu}_{s}}$ we have
\bea
P( \bm{x} \mid \bm{s}, \phi ) 
& = & 
N( (\bm{f}(\bm{\mu}_{s}) - M \bm{\mu}_{s}) + M \bm{s}, \Sigma_{\epsilon})(\bm{x}) \nn \\
&=&
N( M \bm{s}, \Sigma_{\epsilon})(\bm{x-(\bm{f}(\bm{\mu}_{s}) - M \bm{\mu}_{s})}) \nn \\
\eea
so that the appropriate changes to be made to the GKF update equations are simply
\bea
& \bm{x} \rightarrow \bm{x-(\bm{f}(\bm{\mu}_{s}) - M \bm{\mu}_{s})} \nn \\
& M \rightarrow \frac{\partial \bm{f}}{\partial \bm{s}} \mid_{\bm{\mu}_{s}}
\eea
while everything else otherwise remains the same.


\newpage
%
\nocite{*}
\bibliographystyle{plain}
\bibliography{wolf_gkf_arxiv}

\end{document}